%% file: main.tex
\documentclass[letterpaper, 10 pt, conference]{ieeeconf}

\IEEEoverridecommandlockouts                              
\usepackage{times}
\usepackage[ruled,vlined,linesnumbered]{algorithm2e}
\usepackage{multicol}
\usepackage[bookmarks=true]{hyperref}
\usepackage{graphicx}
\usepackage{tabularx}
\usepackage{amssymb}
\usepackage{amsmath}
\usepackage{fancyhdr}
\usepackage{tikz}
\usepackage{bm}
\usepackage{multicol}
\usepackage{listings}
\usepackage{subfigure}
\usepackage{xcolor}
\usepackage{scalerel}
\usepackage[colorinlistoftodos]{todonotes}

\newcommand{\vect}[1]{\mathbf{#1}} 
\newcommand{\vectg}[1]{\boldsymbol{#1}}

\newcommand{\bx}{\mathbf{x}}

\newcommand{\ba}{\mathbf{a}}
\newcommand{\bu}{\mathbf{u}}
\newcommand{\bs}{\mathbf{s}}
\newcommand{\bp}{\mathbf{p}}

\newcommand{\bX}{\boldsymbol{\tau}}
\newcommand{\bU}{\boldsymbol{\xi}}
\newcommand{\bK}{\mathbf{K}}

\newcommand{\bnu}{\boldsymbol{\nu}}

\newcommand{\bbbeta}{\boldsymbol{\beta}}
\newcommand{\bphi}{\boldsymbol{\phi}}
\newcommand{\bpsi}{\boldsymbol{\psi}}
\newcommand{\bl}{\boldsymbol{\ell}}
\newcommand{\bpi}{\boldsymbol{\pi}}
\newcommand{\bmu}{\boldsymbol{\mu}}

\newcommand{\bSigma}{\boldsymbol{\Sigma}}
\newcommand{\bff}{\mathbf{f}}

\DeclareMathOperator*{\argmin}{arg\,min}
\DeclareMathOperator{\atantwo}{atan2}
\DeclareMathOperator{\atan}{atan}

\allowdisplaybreaks
\begin{document}

\begin{titlepage}
\vspace*{\fill}
{\large
\copyright 2025 IEEE.  Personal use of this material is permitted. Permission from IEEE must be obtained for all other uses, in any current or future media, including reprinting/republishing this material for advertising or promotional purposes, creating new collective works, for resale or redistribution to servers or lists, or reuse of any copyrighted component of this work in other works.}
\vspace*{\fill}
\end{titlepage}

\title{Robust Perception-Based Navigation using PAC-NMPC with a Learned Value Function}
\author{Adam Polevoy$^{1, 2}$, Mark Gonzales$^{1, 2}$, Marin Kobilarov$^{2}$, Joseph Moore$^{1, 2}$
\thanks{$^{1}$Johns Hopkins University Applied Physics Lab \newline \hspace*{1.6em}
{\tt\small \{Adam.Polevoy,Mark.Gonzales,\newline \hspace*{1.6em} Joseph.Moore\}@jhuapl.edu} \newline \hspace*{0.8em} $^{2}$Johns Hopkins University Whiting School of Engineering \newline \hspace*{1.6em} {\tt\small mkobila1@jhu.edu}}}
\maketitle

\begin{abstract}

Nonlinear model predictive control (NMPC) is typically restricted to short, finite horizons to limit the computational burden of online optimization. As a result, global planning frameworks are frequently necessary to avoid local minima when using NMPC for navigation in complex environments. By contrast, reinforcement learning (RL) can generate policies that minimize the expected cost over an infinite-horizon and can often avoid local minima, even when operating only on current sensor measurements. However, these learned policies are usually unable to provide performance guarantees (e.g., on collision avoidance), especially when outside of the training distribution. In this paper, we augment Probably Approximately Correct NMPC (PAC-NMPC), a sampling-based stochastic NMPC algorithm capable of providing statistical guarantees of performance and safety, with an approximate perception-based value function trained via RL. We demonstrate in simulation that our algorithm can improve the long-term behavior of PAC-NMPC while outperforming other approaches with regards to safety for both planar car dynamics and more complex, high-dimensional fixed-wing aerial vehicle dynamics. We also demonstrate that, even when our value function is trained in simulation, our algorithm can successfully achieve statistically safe navigation on hardware using a 1/10\textsuperscript{th} scale rally car in cluttered real-world environments using only current sensor information.

\end{abstract}
\IEEEpeerreviewmaketitle

\input{introduction_v2}
\input{related_work}
\input{background_v2}
\input{approach}
\input{sim_experiments}
\input{hardware_experiments}
\input{discussion}

\bibliographystyle{IEEEtran}
\bibliography{references}

\end{document}

%% file: introduction_v2.tex
\section{Introduction}

Nonlinear model predictive control (NMPC) has been an effective approach for perception-based robot navigation (e.g., \cite{ falanga2018pampc, brito2019model, polevoy2022post, polevoy2022complex}).  In some cases, robust and stochastic NMPC (RNMPC and SNMPC respectively) can provide performance guarantees in the presence of uncertainty \cite{lew2021sampling, yin2023risk, 10250934}. However, NMPC is typically restricted to short, finite horizons to limit the computational burden of online optimization. This often limits controller performance, since the cost accrued over the short horizon is often a poor approximation of the true cost-to-go. In navigation tasks, this limitation can cause the system to get caught in local minima. This is especially true in cluttered environments and in ``reactive'' control paradigms that only leverage current sensor measurements. Moreover, hand-designing a terminal cost that incorporates perception data and effectively approximates the true cost-to-go can be prohibitively difficult.


A common approach to improve the global performance of a receding-horizon controller is to utilize a global planner to generate receding-horizon waypoints.
These global planners, however, often do not utilize the true system dynamics, which can be high-dimensional for complex systems. Rather, they tend to leverage simplified approximations with reduced dimensionality to improve computational tractability (e.g., \cite{jian2023dynamic, polevoy2022post, wang2022chase}). There are also challenges associated with implementing an effective global planner for perception-based navigation, such as planning for potential hidden obstacles in obscured areas. In these cases, the performance of the overall system can be limited by the global planner.

By contrast, reinforcement learning (RL) can train effective control policies which directly use current
perception information to avoid local minima that can be observed or inferred in complex environments \cite{singh2022reinforcement, zhu2021deep, dong2023review}. However, because these RL algorithms require large amounts of data, they are often trained in simulation (e.g., \cite{mirowski2016learning, zhu2017target, long2018towards}). This dependence on simulation and the inability of the RL trained policies to guarantee the satisfaction of general state constraints can frequently lead to degraded, and potentially unsafe, performance when the policy is deployed in the real-world or when it is used in out-of-distribution environments. 

\begin{figure}[]
    \centering
    \includegraphics[trim={0 5 0 20},clip,width=0.9\columnwidth]{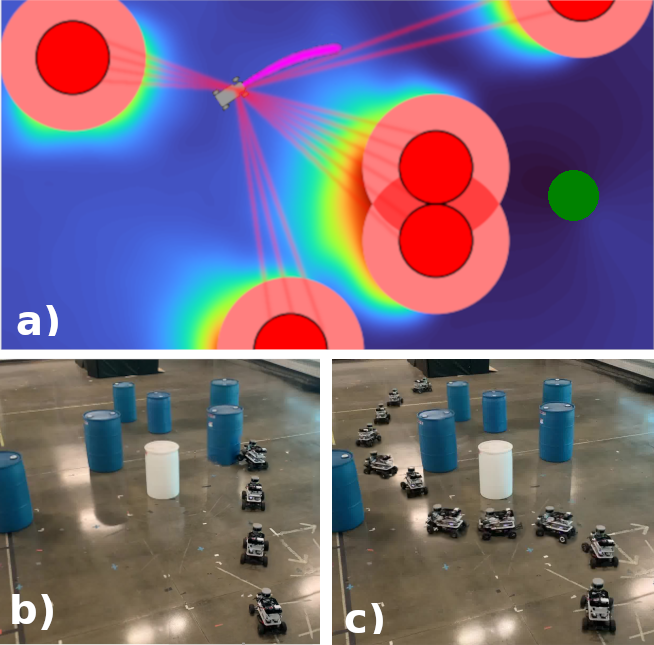}
    \caption{a) Rally car navigating through obstacle field using NMPC with learned value function and LiDAR. b) Timelapse of rally car navigating through barrels using RL trained policy. c) Timelapse of rally car navigating through barrels using PAC-NMPC with learned value function.}
    \vspace{-1.5em} 
    \label{fig:main_img}
\end{figure}

In this paper, we augment Probably Approximately Correct NMPC (PAC-NMPC) \cite{10250934} with an RL-trained perception-based value function to achieve navigation with statistical performance guarantees even when the learned value function is approximate and uncertain. Using Monte Carlo dropout to capture network uncertainty, we learn a stochastic model of the value function. We then use PAC-NMPC to minimize a bound on the expected terminal costs and constraints derived from this value function. This encourages our receding-horizon NMPC algorithm to assume the long-range behavior of the RL policy, while preserving the statistical safety guarantees afforded by minimizing the sample-complexity bounds derived in \cite{kobilarov2015sample}. We demonstrate through simulation and hardware experiments that our algorithm can achieve probabilistically safe perception-based navigation, improved sim-to-real transfer, robustness to out-of-distribution scenarios, and scales to complex, nonlinear high-dimensional dynamic systems.

%% file: related_work.tex
\section{Related Work}

To enable effective finite horizon NMPC, researchers have investigated the use of learned models to generate waypoints directly from sensor data \cite{sharma2012autonomous}. Learned waypoints have been used in MPC for drone racing \cite{kaufmann2018deep}, quadrotor navigation in environments with dead-end corridors \cite{greatwood2019reinforcement}, navigation in real-world cluttered environments \cite{bansal2020combining}, and navigation in dynamic environments with other agents \cite{brito2021go}. Recently, learned waypoints have been used alongside artificial potential fields \cite{bektacs2022apf} and hierarchical reinforcement learning \cite{gao2023efficient} to improve navigation in complex environments when using only immediate sensor information. 

By contrast, our approach aims to use reinforcement learning to augment the cost function of NMPC directly, which is more closely related to Quasi-Infinite Horizon NMPC \cite{chen1998quasi}. A more recent approach suggested the use of a learned approximate value function \cite{zhong2013value} to determine the terminal cost. This idea was built upon to update the approximate value function online \cite{lowrey2018plan}, learn an approximate dynamics model \cite{hong2019model}, formulate the running cost as an importance sampler of the value function \cite{hoeller2020deep}, and to weigh the value function against local Q-function approximations along the trajectories \cite{bhardwaj2020blending}. 


Safe RL methods aim to generate a policy that can minimize a cost while satisfying safety constraints \cite{gu2022review, brunke2022safe}. Often these problems are posed as Constrained Markov Decision Processes (CMDP) and are solved using Lagrangian methods \cite{geibel2005risk,altman1998constrained}, or via constrained policy gradient methods \cite{achiam2017constrained}. Other methods, such as Conservative Safety Critic \cite{bharadhwaj2020conservative}, learn a safety critic, which is used to select safe actions to execute. Safety Editor Policy \cite{yu2022towards} satisfies constraints by learning an additional policy that maps unsafe actions to safe actions. Control barrier functions  (CBF) (e.g., \cite{anand2021safe}), Lyapunov functions (e.g., \cite{chow2018lyapunov}, and backwards reachable sets (e.g., \cite{fisac2019bridging}) are also used for ensuring safety in terms of RL constraint satisfaction. RL-CBF \cite{cheng2019end} learns a Gaussian Process model for the unknown dynamics to generate a CBF, which is then used as a safety filter on the RL-policy and to improve exploration efficiency. However, this approach assumes the algorithm is provided with a valid safe set, which can be very challenging to compute. Another approach, \cite{marvi2021safe}, adds a CBF to the value function and proves that the learned actor and critic will converge to a safe, optimal solution. However, the CBF must be known and included in the augmented reward function. Some of these model-based safe RL approaches have been demonstrated in hardware (e.g., CBF methods for autonomous driving in \cite{ma2021model} and reachable set methods for legged locomotion \cite{yang2022safe}). Similar to our method, Actor Critic Props \cite{sheckells2019actor}, uses sample complexity bounds to provide probabilistic guarantees on the performance and safety of learned policies. Generally, the guarantees produced by this method do not hold outside of the training distribution. Safe RL has also been formulated using robust MPC, rather than neural networks, as the function approximator \cite{zanon2020safe, gros2022learning}. Although this inherently generates policies that satisfy constraints, it is limited by the parameterization of the MPC policy.

To the best of our knowledge, our approach is the first to combine a learned perception-based stochastic value function with NMPC to achieve probabilistic guarantees for navigation and obstacle avoidance.

%% file: background_v2.tex
\section{Background}


\subsection{PAC-NMPC}
Probably Approximately Correct NMPC (PAC-NMPC) \cite{10250934} is a sampling-based SNMPC method which minimizes an upper confidence bound on the expected cost and probability of constraint violation of a local feedback policy. 

Consider the stochastic dynamical system given by the probability density function $p(\bx_{t+1}|\bx_{t}, \bu_{t})$, where $\bx_{t}\in\mathbb{R}^{N_x}$ is a vector of state values and $\bu_{t}\in\mathbb{R}^{N_u}$ is a vector of control inputs. PAC-NMPC uses Iterative Stochastic Policy Optimization (ISPO) \cite{kobilarov2015sample} to formulate the search for a control policy, $\bu_{t}= \bpi(\bx_t,\bU)$, as a stochastic optimization problem. This is achieved by sampling the policy parameters, $\bU$, from a surrogate distribution given by the probability density function $p(\bU|\bnu)$ and defined by hyper-parameters $\bnu$.

More specifically, PAC-NMPC optimizes a time-varying local feedback policy of the form $\bu_t =\bK_t(\bX^d(\bU,\bx_0))\left(\bx_t^d(\bU,\bx_0) - \bx_t\right) + \bu_t^d(\bU)$ where $\bX^d\triangleq\{\bx_0^d,\bu_0^d, \bx_1^d,\bu_1^d , ... , \bu_{N_T-1}^d, \bx_{N_T}^d\}$ is the nominal trajectory computed using a nominal deterministic discrete-time dynamics model
$\bx_{t+1}^d=\bx_t^d+\bff(\bx_t^d,\bu_t^d)\Delta t$
over $N_T$ time steps.
Here $\bx_t^d\in \mathbb{R}^{N_x}$ and $\bu_t^d\in \mathbb{R}^{N_u}$ are the nominal states and control inputs at time index $t$ respectively. $\Delta t$ is the discrete time step and $\bK_t(\bX^d(\bU,\bx_0))\in\mathbb{R}^{N_u\times N_x}$ is a sequence of time-varying feedback gains computed using the finite horizon, discrete, time-varying linear quadratic regulator (TVLQR). The policy is parameterized only by the nominal input sequence, so that $\bU ~= [ {\bu^d_0}^T \ {\bu^d_1}^T \ ... \ {\bu^d_{N_T-1}}^T]^T$. The surrogate distribution,  $p(\bU|\bnu)$, is parameterized as a multivariate Gaussian, $\mathcal{N}\left(\bU | \bmu, \bSigma \right)$, with a mean, $\bmu$, and a covariance matrix, $\bSigma$. Since the covariance matrix is diagonal, the hyperparameters can be written as $\bnu \triangleq [\bmu^T \ diag(\bSigma)^T]^T$.

For a discrete-time trajectory sequence $\bX=\{\bx_0,\bu_0, \bx_1,\bu_1 ..., \bu_{N_T-1}, \bx_{N_T} \}$, one can define a non-negative trajectory cost function, $J(\bX) \ge 0$, and a trajectory constraint violation function $C(\bX) \in \{0, 1\}$. During each planning interval, PAC-NMPC iteratively optimizes $\bnu^* = \argmin_{\bnu}\min_{\alpha>0} (\mathcal{J}^+_\alpha(\bnu) + \gamma \mathcal{C}^+_\alpha(\bnu))$, where $\mathcal{J}^+_\alpha(\bnu)$ is the PAC bound on $J(\bX)$, $\mathcal{C}^+_\alpha(\bnu)$ is the PAC bound on $C(\bX)$, and $\gamma > 0$ is a  heuristically selected weighting coefficient. $\mathcal{J}^+_\alpha(\bnu)$ and $\mathcal{C}^+_\alpha(\bnu)$ take the form
\begin{align}
&\mathcal{J}^+_\alpha(\vectg{\nu})\! \triangleq \! \widehat{\mathcal J}_\alpha(\vectg{\nu}) + \alpha d(\vectg{\nu}) + \Phi_{\alpha}(\delta), 
\label{eq:pacbound} \\
&\widehat {\mathcal J}_\alpha(\bnu) \triangleq \frac{1}{\alpha LM} \sum_{i=0}^{L-1} \sum_{j=1}^{M} \zeta\left(\alpha \ell_{ij} \right), \notag\\
&\zeta(x) = \log\left(1+x+\frac{1}{2}x^2\right), 
\ell_{ij} = J(\bX_{ij})\frac{p(\bU_{ij}|\bnu)}{p(\bU_{ij}|\bnu_i)},
\notag\\
 & d(\bnu)\! \triangleq \! \frac{1}{2L}\sum_{i=0}^{L-1}b_i^2e^{D_2\left(p(\cdot | \bnu)||(p(\cdot | \bnu_i)\right)}, \Phi_{\alpha}(\delta)\! \triangleq \frac{1}{\alpha LM}\log\frac{1}{\delta},\notag\\
&  0 \leq J(\bX_{ij}) \leq b_i \; \forall j = 0, ..., M \notag
\end{align}
where $\widehat{\mathcal J}_\alpha(\vectg{\nu})$ is a robust estimator of the expected cost, $d(\vectg{\nu})$ is a distance between distributions, and $\Phi_{\alpha}(\delta)$ is a concentration-of-measure term, $\alpha > 0$ is an annealing coefficient, $1 - \delta$ is the bound confidence, $L$ is the number of prior policies, $M$ is the number of samples per policy, and $D_2$ is the Renyi divergence.

These PAC bounds were derived in \cite{kobilarov2015sample} and guarantee that $\mathbb{P}\left(\mathbb{E}_{\bX, \bU \sim p(\cdot, \cdot|\bnu)}\left[J(\bX)\right] \leq \mathcal{J}^+_\alpha(\bnu)\right) \geq 1 - \delta$ and $  \mathbb{P}\left(\mathbb{E}_{\bX, \bU \sim p(\cdot, \cdot|\bnu)}\left[C(\bX)\right] \leq \mathcal{C}^+_\alpha(\bnu)\right) \geq 1 - \delta \label{eq:constraint_pac_bound}$. They are not only optimization targets, but also provide probabilistic guarantees of performance and safety. For instance, setting $\delta=0.05$ means that with 95\% confidence the expected cost will not exceed $\mathcal J_{\alpha}^+$, while the probability of constrained violation will not exceed $\mathcal C_{\alpha}^+$. 

\subsection{Actor Critic Reinforcement Learning}
\label{sec:ACRL}
Actor Critic algorithms are policy gradient approaches to solve Markov decision processes (MDP) by simultaneously learning a policy and action-value function \cite{sutton2018reinforcement}. Consider a MDP with states $\bs_t \in \mathbb{R}^{N_s}$, actions $\ba_t \in \mathbb{R}^{N_a}$, transition dynamics $p(\bs_{t+1}|\bs_{t}, \ba_{t})$, and reward function $r(\bs_t, \ba_t)$, where $t$ denotes the time index. The discounted return is defined as the sum of the discounted future rewards of a state-action sequence, $R^{\gamma_d}_t = \sum_{i=t}^\infty\left\{\gamma_d^{i-t}r(\bs_i, \ba_i)\right\}$ where, $\gamma_d \in [0, 1)$ is a discount factor. The action-value function (Q-function), $Q^{\bpi}(\bs_t, \ba_t) = \mathbb{E}\left[R^{\gamma_d}_t \middle| \bs_t, \ba_t\right]$, is the expected discounted return after taking an action $\ba_t$ from state $\bs_t$ and then following policy $\ba_t = \bpi(\bs_t)$. The value function, $V^{\bpi}(\bs_t) = Q^{\bpi}(\bs_t, \bpi(\bs_t))$, is the expected discounted return of following a policy, $\ba_t = \bpi(\bs_t)$, from a given state $\bs_t$. Actor-critic algorithms learn parameters $\bphi$ for an actor policy, $\bpi^{\bphi}(\bs_t)$, and parameters $\bpsi$ for a critic action-value function, $Q^{\bpsi}(\bs_t, \ba_t)$, which estimate the optimal policy $\bpi^*(\bs_t)$ and action-value function $Q^{\bpi^*}(\bs_t, \ba_t)$ where $\bpi^*(\bs_t) = \ba_t^* = \argmin_{\ba_t} \mathbb{E}\left[Q^{\bpi^*}(\bs_t, \ba_t)\right]$ and $Q^{\bpi^*}(\bs_t, \ba_t^*) = \mathbb{E}\left[{r(\bs_t, \ba^*_t) + \gamma_d Q^{\bpi^*}(\bs_{t+1}, \ba_{t+1}^*)}\right]$.

%% file: approach.tex
\section{Approach}


\subsection{Problem Formulation}

Our objective is for a robot to navigate through an obstacle field from an initial state, $\bx_I$, to a goal state, $\bx_G$. The robot is equipped with LiDAR which returns range measurements $\bl_t = [\ell_t^0 \ \ell_t^1 \ ... \ \ell_t^{N_{\bl}}]$ at bearings $\bbbeta = [\beta^0 \ \beta^1 \ ... \ \beta^{N_{\bl}}]$ with a maximum range of $\ell_{max}$ at each timestep. We formulate a generic cost on a finite horizon trajectory,
\begin{align}
    J(\bX) = \sum\nolimits_{i=0}^{N_T-1}\left\{q(\bx_i, \bu_i)\right\} + q_f(\bx_{N_T}),
\end{align}
where $q(\bx_i, \bu_i)$ is the running cost and $q_f(\bx_{N_T})$ is the terminal cost. We also formulate constraints to bound the state and prevent collisions with obstacles:
\begin{align}
g_b(\bx_t) &= (\bx_t - \bx_l < 0) \lor (\bx_u - \bx_t < 0),  \\
g_{o}(\bx_t) &= \neg \left((dist(\bx_t, \bp_{o^j}) > r) \ \forall \ j\right), \nonumber\\
c(\bx_t) &= g_b(\bx_t) \lor g_{o}(\bx_t), \nonumber\\
C(\bX) &= c(\bx_0) \lor c(\bx_1) \lor \cdots \lor c(\bx_{N_T})\nonumber.
\end{align}


Here, $\lor$ and $\neg$ are the logical or and not operators, respectively. $\bx_l$ is the lower state bound, $\bx_u$ is the upper state bound, $r$ is the maximum robot radius.  $\bp_{o^j}$ is an observed occupied point in the world frame, calculated as
\begin{align}
    \bp_{o^j} = \bp_t + \begin{bmatrix}\cos(\theta_t + \beta^j) & \sin(\theta_t + \beta^j)\end{bmatrix}^T\ell_t^j
    \label{eq:obstacles}
\end{align}
for each $\ell_t^j < \ell_{max}$ where $\bp_t$ and $\theta_t$ are the position and orientation of the robot at state $\bx_t$.

\subsection{Learned Value Function}
Using our probabilistic dynamics model,  sensor model, costs and constraints, we define the components of the general MDP referenced in Section \ref{sec:ACRL}. Let $\bs_t = {\bf h}(\bx_t, \bl_t)$, $\ba_t = \bu_t,$ and
\begin{align}
r(\bs_t, \ba_t) = -q(\bx_t, \bu_t)-\gamma_r c(\bx_t),
\label{eq:reward_function}
\end{align}
where ${\bf h}(\bx_t, \bl_t)$ maps the robot state and sensor measurements to the MDP state representation and $\gamma_r$ is a heuristically selected constraint violation penalty. Although this problem is actually a partially observable MDP (POMDP), we approximate it as an MDP by including the LiDAR measurements as a part of the state vector, which is a common approach \cite{tai2017virtual, de2021soft, beomsoo2021mobile}.

We learn an actor policy, $\bpi^{\bphi}(\bs_t)$, and a critic action-value function, $Q^{\bpsi}(\bs_t, \ba_t)$, using the CleanRL implementation \cite{huang2022cleanrl} of TD3 \cite{fujimoto2018addressing}, a state-of-the-art actor-critic method. However, our approach is agnostic to the method used to train the actor and critic networks. The value function is then reconstructed as $V^{\bpsi\bphi}(\bx_t, \bl_t) = Q^{\bpsi}(\bs_t, \bpi^{\bphi}(\bs_t))$.

\subsection{PAC-NMPC with Learned Value Function}

To use the learned value function as a terminal state cost, we must estimate the future LiDAR measurement, $\widehat{\bl}_{N_T}$, at the terminal state based on the current LiDAR measurement, $\bl_0$. 
The performance guarantees on the value function assume that this future sensor prediction is accurate. We calculate the range, $\widehat{\ell}^j$, and bearing, $\widehat{\beta}^j$, to each observed occupied point from the terminal state of the trajectory as $\widehat{\ell}^j = \| \bp_{N_T} - \bp_{o^j} \|$, and $\widehat{\beta}^j = \atantwo(\bp_{N_T} - \bp_{o^j}) - \theta_{N_T}$.
Since the bearing estimates do not necessarily correspond to the discrete bearing values used by the sensor, we assign the the estimated LiDAR measurements to the closest bearing $\widehat{\ell}_{N_T}^k = \widehat{\ell}^j$ where
$k = \argmin_{k}\{|\widehat{\beta}^j-\beta^k|\}$.



We use the negative learned value function as the terminal cost so that $q_f(\bx_{N_T}) = -V^{\bpsi\bphi}(\bx_{N_T}, \widehat{\bl}_{N_T})$. In this way, we approximate the infinite horizon trajectory cost given the current LiDAR measurement. 
We apply the additional constraint that the learned value function improves from the current state to the final state of the trajectory:
\begin{align}
    g_{V}(\bx_{N_T}) &= V(\bx_{N_T}, \widehat{\bl}_{N_T}) - V(\bx_0, \bl_0) \\
    c_{V}(\bx_{N_T}) &= g_{V}(\bx_{N_T}) < 0 \nonumber\\
    C(\bX) &= c(\bx_0) \lor c(\bx_1) \cdots \lor c(\bx_{N_T}) \lor c_{V}(\bx_{N_T})\nonumber. \label{eq:value_function_constraint}
\end{align}
During each planning interval, PAC-NMPC optimizes and returns policy hyperparameters, $\bnu^*$, a PAC bound on the expected cost, $\mathcal{J}^+_\alpha(\bnu^*)$, and a PAC bound on the probability of constraint violation, $\mathcal{C}^+_\alpha(\bnu^*)$. These bounds serve as performance and safety guarantees for the policy; the probability of violating the obstacle constraint or worsening the learned value function, given the current LiDAR measurement, will be less than $\mathcal{C}^+_\alpha(\bnu)$ with a probability of $1-\delta$ (eq. \ref{eq:constraint_pac_bound}).

%% file: sim_experiments.tex
\section{Simulation Experiments}

\subsection{Experimental Setup}

We simulate a stochastic bicycle model with acceleration and steering rate inputs given as
\begin{align}
  \bx_{t+1} &\sim p(\cdot | \bx_t, \bu_t) \triangleq \bx_t + \left(f(\bx_t, \bu_t) + \boldsymbol{\omega} \right) \Delta t, \nonumber \\
  f(\bx_t, \bu_t) &= [v\cos(\theta), v\sin(\theta), v\tan(\delta_s) / l, \dot{v}, \dot{\delta_s}]^T, \nonumber \\
\vectg{\omega} &\sim \mathcal{N}(\cdot | \mathbf{0}, \boldsymbol{\Gamma}).
\end{align}
Here ${\bx_t}=[p_x \ p_y \ \theta \ v \ \delta_s]^T$ is the state vector, ${\bu_t}=[\dot{v} \ \dot{\delta_s}]^T$ is the control vector, $l = 0.33 m$ is the wheelbase, $\boldsymbol{\omega}$ is Gaussian noise, and $\boldsymbol{\Gamma} = diag(\left[4{\rm e}^{-4}, 4{\rm e}^{-4}, 1.1{\rm e}^{-2}, 1{\rm e}^{-1}, 5.6{\rm e}^{-3}\right])$ is the process covariance, which was fit from hardware data. We limit the acceleration, $\dot{v}\in\left[-1., 1\right] \frac{m}{s}$, the steering rate, $\dot{\delta_s}\in\left[-1, 1 \right] \frac{rad}{s}$, and the steering angle, $\delta_s\in\left[-0.4, 0.4\right] rad$.

We use a quadratic state cost $q(\bx, \bu) = (\bx-\bx_G)^T Q (\bx-\bx_G)$ where $Q = diag([1{\rm e}^{-2} \ 1{\rm e}^{-2} \ 0 \ 0 \ 0])$, and a velocity constraint such that $v \in \left[-1, 3\right] \frac{m}{s}$.

Our approach considers not only uncertainty in the dynamics, but also in the learned actor and critic networks. For every trajectory sampled by PAC-NMPC, we also sample dropout masks for the actor and critic networks \cite{gal2016dropout} before evaluating the cost-to-go at the trajectory's terminal state. Since our approach already requires multiple forward passes through the networks, one for each trajectory, Monte Carlo dropout does not add additional inference time. This allows us to incorporate value function uncertainty into the PAC bound to provide more accurate performance guarantees.

We compared our approach against four baselines across two simulation environment distributions. The actor and critic networks were trained for each environment. First, we evaluated PAC-NMPC with a quadratic terminal cost, $q_f(\bx_{N_T}) = (\bx_{N_T}-\bx_G)^T Q_f (\bx_{N_T}-\bx_G)$ where $Q_f = diag([1 \ 1 \ 0 \ 0 \ 0])$. Second, we used the LiDAR measurements to continuously build and maintain an occupancy grid of the environment. Each planning iteration, we ran a naive A* search algorithm on the grid to find the shortest path to the goal. A receding horizon goal, $\bx_A$, was selected at distance of $v_{max} \cdot N_t \cdot \Delta t$ along the path, and was used to form a quadratic terminal cost, $q_f(\bx_{N_T}) = (\bx_{N_T}-\bx_A)^T Q_f (\bx_{N_T}-\bx_A)$. Third, we compared directly against the actor policy, $\bpi^{\bphi}(\bs)$. Fourth, instead of using PAC-NMPC, we optimized trajectories with MPPI \cite{7487277}, while still using the learned value function as a terminal trajectory cost.

When running PAC-NMPC, we optimize a 12 timestep trajectory with $\Delta t=0.1$ sec at a replanning period of $H=0.2$ sec, and we interpolate the feedback policies to 50Hz. We set PAC-NMPC parameters as $L=5$, $M=1024$, $\delta=0.05$, and normalize the trajectory costs before optimization to achieve tighter PAC bounds. We evaluated PAC-NMPC with several $\gamma$ values and selected the best value for each environment; $\gamma=2$ in the cluttered environments and $\gamma=4$ in the concave trap environments. When running MPPI, we similarly normalized the sampled trajectory costs and summed them with the constraints using the same $\gamma$ values. We used the same number of timesteps, $\Delta t$, $H$, number of sampled trajectories, a temperature of $\gamma_t=0.35$, and a sampling variance of $\Sigma_\epsilon=0.01$. We allowed MPPI to run for as many iterations as possible in the replanning period and interpolated the resulting trajectories to 50Hz.

\begin{figure}[t]
    \vspace*{0.75em} 
    \centering
    \includegraphics[trim={45 0 80 0},clip,width=\columnwidth]{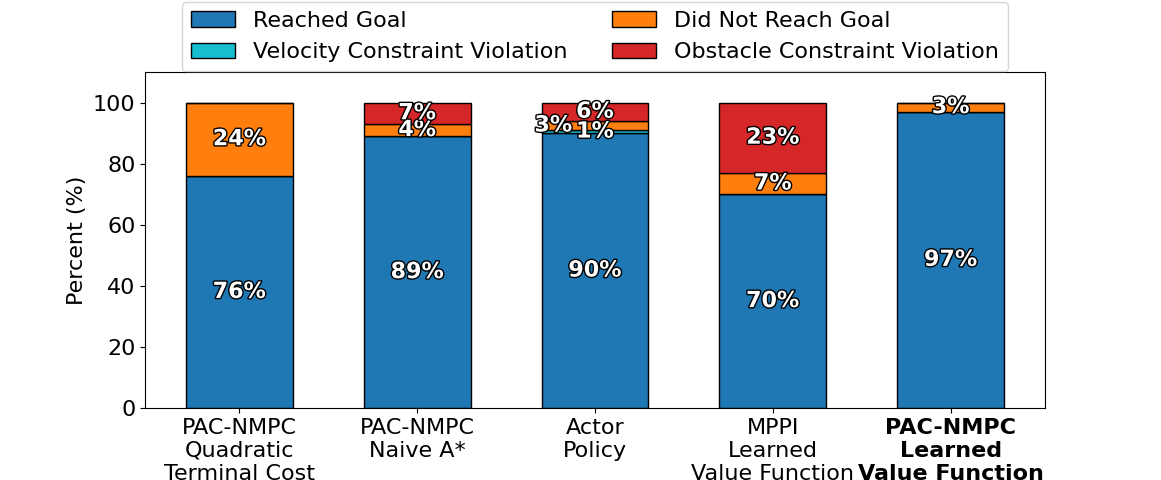}
    \caption{Simulated cluttered environments. Percentage of trials in which the system reached the goal, didn't reach the goal, and violated constraints.}
    \label{fig:sim_rates}
\end{figure}

\begin{figure}[t]
    \centering
    \includegraphics[trim={70 20 85 55},clip,width=0.78\columnwidth]{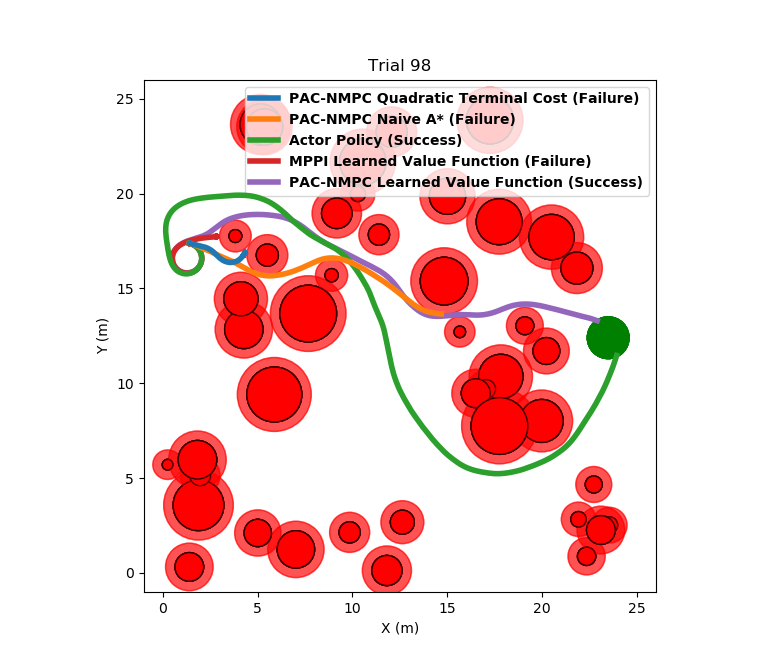}
    \caption{Simulated cluttered environments. Testing environment where PAC-NMPC fails when only using a quadratic terminal cost.}
    \label{fig:pacnmpc_fail}
    \vspace{-1.5em} 
\end{figure}

We trained the actor and critic in simulation with the stochastic bicycle model and a 64 beam $360^\circ$ LiDAR. The MDP state, $\bs_t$, consists of the normalized velocity, normalized tangent of the steering angle, normalized range to the goal, cosine and sine of the bearing to the goal, and normalized LiDAR measurement. The reward directly mirrors the NMPC costs and constraints (Eq. \ref{eq:reward_function}) with $\gamma_r = 1000$. We used shallow, fully connected neural networks as the function approximators for $\bpi^{\bphi}(\bs_t)$ and $Q^{\bpsi}(\bs_t, \ba_t)$. Simulations were run on a laptop with an Intel Core i-9-13900H CPU and a Nvidia GeForce RTX 4080 Max-Q GPU.

\subsection{Cluttered Environments}

We generated 100 random environments from the training distribution. Environments consisted of an initial state, $\bx_I$, a goal state $\bx_G$, and circular obstacles. The obstacles were allowed to overlap, which allowed the constraint regions to combine to form complex environments. To provide meaningful testing environments, environments in which no obstacles blocked the path to the goal were discarded.

\begin{figure}[t]
    \vspace*{0.75em} 
    \centering
    \includegraphics[trim={70 10 90 80},clip,width=0.9\columnwidth]{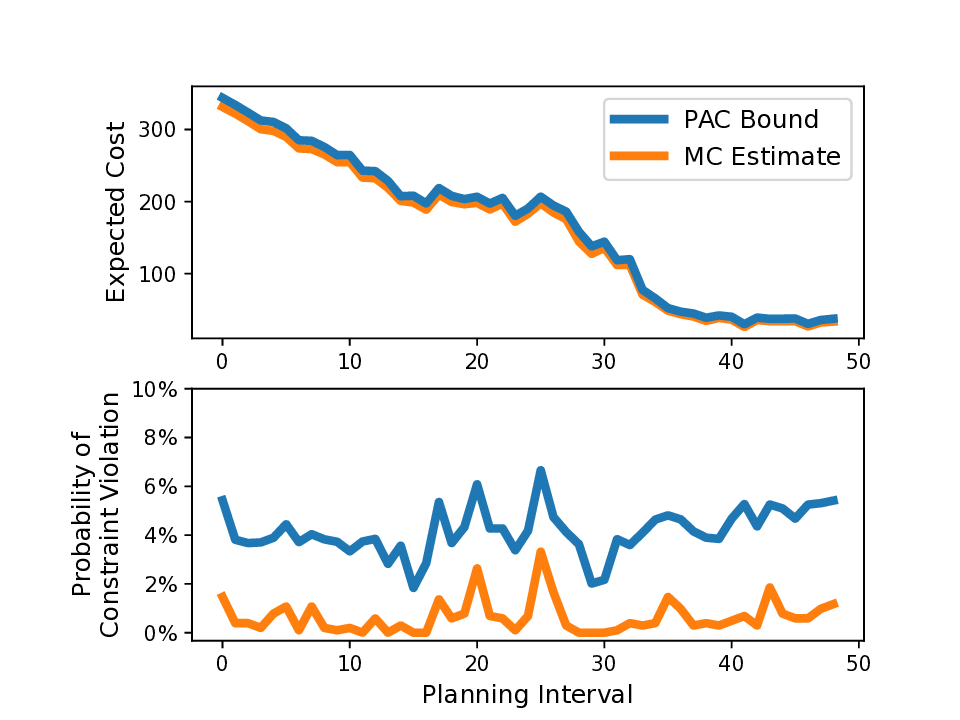}
    \caption{Optimized PAC Bounds compared to Monte Carlo estimates at each planning interval for example trial when using the learned value function.}
    \label{fig:pacbounds}
    \vspace{-1.5em} 
\end{figure}

Results are shown in Figure \ref{fig:sim_rates}, with an example environment in Figure \ref{fig:pacnmpc_fail}. When using a quadratic terminal cost, the system often got caught in local minima. When using A*, the system occasionally planned paths through obscured obstacles, which caused constraint violations if the LiDAR was unable to view the obscured obstacles until too close to recover. The actor policy was unable to explicitly enforce constraints which resulted in violations. Our approach outperformed all baselines and never violated the constraints.

In Figure \ref{fig:pacbounds}, we plot the optimized PAC bounds at each planning interval for an example trial when using the  learned value function as a terminal cost and constraint. To validate PAC bounds, we compare them against Monte Carlo estimates of the expected cost and probability of constraint violation, which were formed by sampling 1024 trajectories and dropout masks. On average, PAC-NMPC produced guarantees that the probability of constraint violation of the NMPC generated policies would be less than 5\%. 

To demonstrate that our approach incorporates uncertainty from the actor and critic networks into the PAC bound computation, we compared the percentage of bound violations that occurred when optimizing the bounds with and without sampling dropout masks. In both cases, we compare against Monte Carlo estimates using 1024 sampled trajectories and dropout masks. We found that across all trials, when optimizing with sampled dropout masks, the expected cost bound was never violated, and the probability of constraint violation bound was violated in only 0.34\% of planning intervals. When optimizing without sampled dropout masks, the expected cost bound was violated in 65.49\% of planning intervals, and the probability of constraint violation bound was violated in 1.45\%.

\subsection{Concave Trap Environments}

\begin{figure}[t]
\vspace*{0.75em} 
\centering
\includegraphics[trim={45 0 80 0},clip,width=\columnwidth]{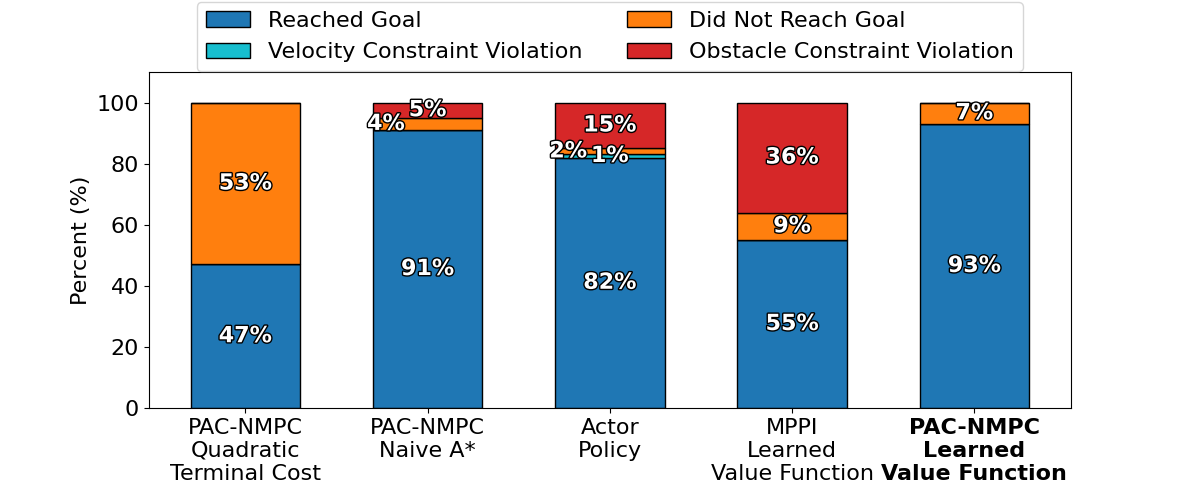}
\caption{Simulated concave trap environments. Percentage of concave trap trials in which the system reached the goal, didn't reach the goal, and violated constraints.}
\label{fig:trap_bar_plot}
\end{figure}

\begin{figure}[t]
\centering
\includegraphics[trim={145 20 160 50},clip,width=0.78\columnwidth]{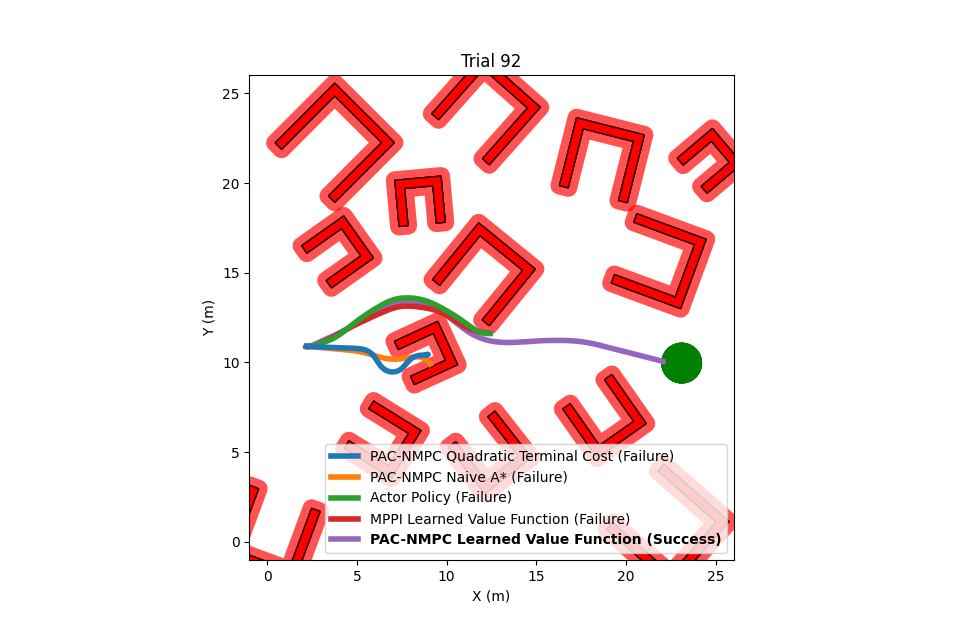}
\caption{Simulated concave trap environments. Concave trap environment where the actor policy fails and PAC-NMPC fails with both the quadratic terminal cost and with A*.}
\label{fig:trap_fails}
\vspace{-1.5em} 
\end{figure}

We also evaluated our approach in environments constructed solely of concave traps to highlight its ability to safely avoid local minima. We uniformly sampled the number of obstacles between 0 and 20, the side lengths of each obstacle between 2.5 and 5.0 meters, and the obstacles were not allowed to overlap. The angles of the obstacles were sampled uniformly such that they were pointing towards the starting position of the robot $\pm\frac{\pi}{2}$ radians. An example environment is shown in Figure \ref{fig:trap_fails}. We generated 100 random environments from the training distribution to evaluate our approach. Our approach outperformed all baselines and never violated the constraints (Fig. \ref{fig:trap_bar_plot})

\subsection{Fixed-wing UAV}

\begin{figure}[t]
    \vspace*{0.75em} 
    \centering
    \includegraphics[trim={0 20 0 40},clip,width=0.75\columnwidth]{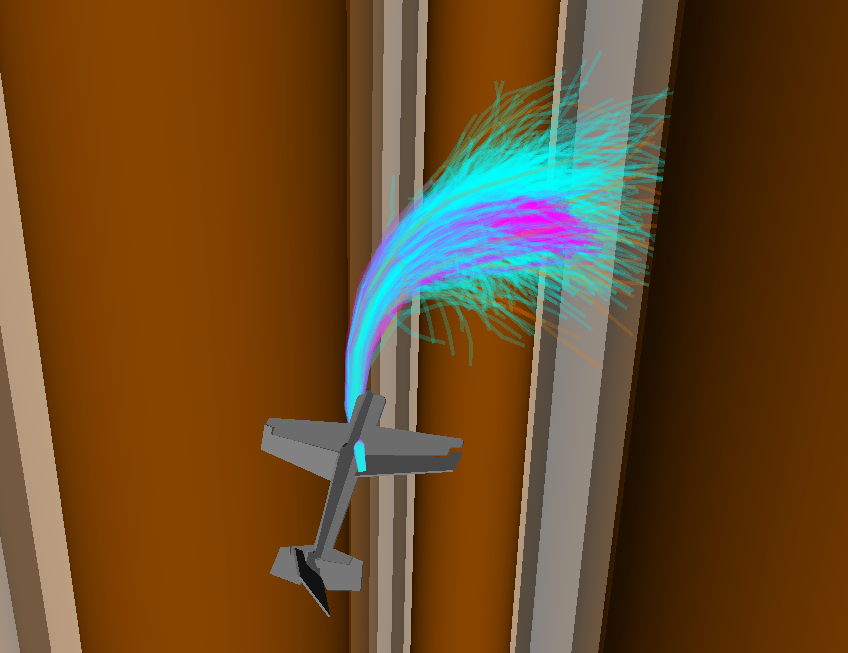}
    \caption{Fixed-wing simulation environment.}
    \label{fig:fixedwing_sim}
    \vspace{-1.5em} 
\end{figure}


We repeat our approach with stochastic fixed-wing UAV dynamics to demonstrate that it can extend to high-dimensional systems (Fig. \ref{fig:fixedwing_sim}). We utilize the formulation of the fixed-wing described in \cite{basescu2020direct} with 2nd order Runge-Kutta (RK2) integration. The state of the system is $\vect{x} = \left[\vect{r}, \vect{q}, \vectg{\delta}, \vect{v}, \vectg{\omega}, p\right]$ where $\vect{r} \in \mathbb{R}^3$ is the position, $\vect{q} \in \mathbb{Q}$ is the orientation, $\vectg{\delta} \in \mathbb{R}^3$ are control surface deflections, $\vect{v} \in \mathbb{R}^3$ is the linear velocity, $\vectg{\omega} \in \mathbb{R}^3$ is the angular velocity, and $p$ is the propeller speed. The control signal is the rate of change of the control surface deflections and of the propeller speed: $\vect{u} = \left[u_a, u_e, u_r, u_p \right]$. We place Gaussian noise over the body frame accelerations, with means of $[0.10, -0.90, -0.89]\frac{m}{s^2}$, $[1.18, -0.16, 1.71]\frac{rad}{s^2}$ and variances of $diag([1.71, 1.63, 1.41])$, $diag([10.05, 5.57, 15.23])$, which was fit from data collected on hardware. We simulated the fixed-wing with a 64 beam 360$^\circ$ LiDAR attached on a gimbal such that it remains parallel to the xy-plane.

\begin{figure}[t]
    \vspace*{0.75em} 
    \centering
    \includegraphics[trim={40 0 75 0},clip,width=\columnwidth]{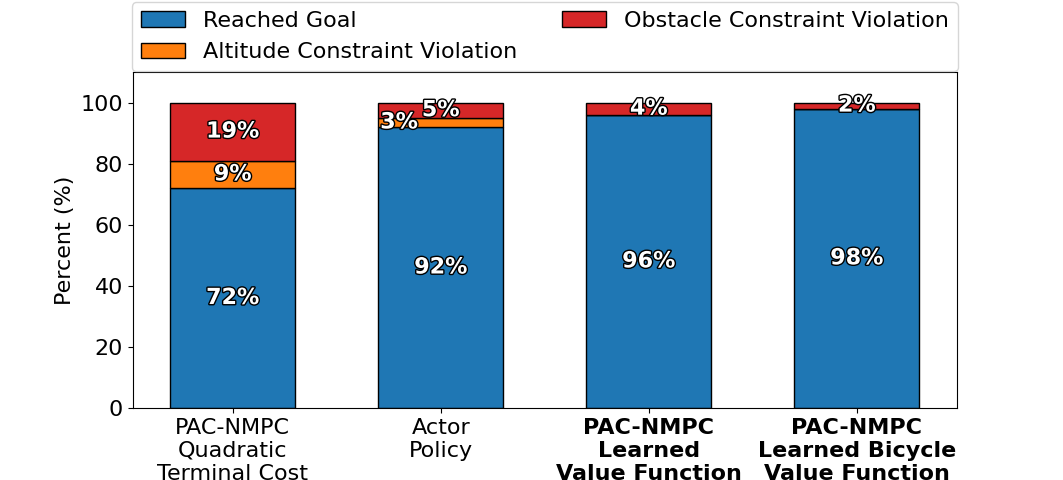}
    \caption{Fixed-wing environments. Percentage of trials in which the system reached the goal or violated constraints.}
    \label{fig:fixedwing}
\end{figure}

\begin{figure}[t]
    \centering
    \includegraphics[trim={40 0 75 0},clip,width=\columnwidth]{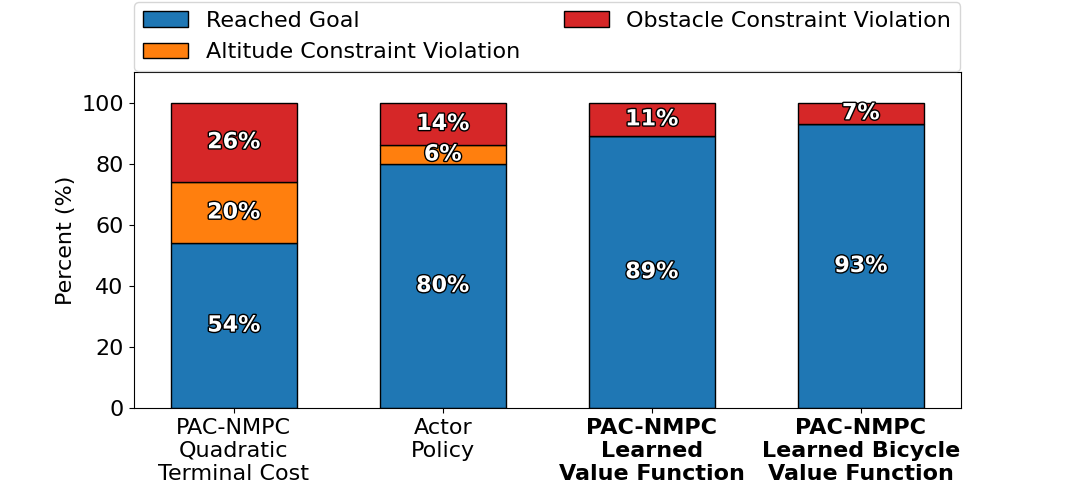}
    \caption{Fixed-wing environments with additional noise. Percentage of trials in which the system reached the goal or violated constraints.}
    \label{fig:fixedwing_noisy}
    \vspace{-1.5em} 
\end{figure}

We use a quadratic state cost $q(\bx, \bu) = (\vect{r}-\bx_G)^T Q_\vect{r} (\vect{r}-\bx_G) + \vectg{\delta}^T Q_{\vectg{\delta}} \vectg{\delta} + \vectg{\omega}^T Q_{\vectg{\omega}} \vectg{\omega}$ where $Q_{\vect{r}} = Q_{\vectg{\omega}} = diag([0.01, 0.01, 0.01])$ and $Q_{\vectg{\delta}} = diag([0.1, 0.1, 0.1])$. We constrained the altitude to $[0, 5] m$, the speed along the $x, y$ plane to $\leq 8 \frac{m}{s}$, and the roll rate to $[-10, 10] \frac{rad}{s}$. We set $L = 1$, $\gamma=3$, and sampled cluttered environments with a maximum of 20 obstacles. The actor and critic networks trained in 3 million steps over 4.5 hours. In these experiments, we  do not count velocity constraint violations in Figures \ref{fig:fixedwing}, \ref{fig:fixedwing_noisy}. Our approach outperformed both the actor policy and PAC-NMPC with a quadratic terminal cost of $q_f(\bx, \bu) = 10.0 \cdot q(\bx, \bu)$ (Fig. \ref{fig:fixedwing}). 

To demonstrate that a value function trained on simple dynamics can aid in controlling more complex systems, we ran experiments utilizing the bicycle value function to control the fixed-wing. We trained the bicycle value function in these environments with $v_{min}=2$, $v_{max}=8$, and no Gaussian noise. Then, we mapped the fixed-wing state to the bicycle state by extracting $p_x$, $p_y$, $\theta$, $v$ from $\vect{r}$, $\vect{q}$, $\vect{v}$ and setting $\delta=\atan(\frac{L}{v}\dot{\theta})$. For the terminal cost, we applied the bicycle value function plus $q(\bx, \bu)$ where $Q_{\vect{r}} = diag([0.00, 0.00, 0.1])$, $Q_{\vectg{\omega}} = diag([0.1, 0.1, 0.1])$ and $Q_{\vectg{\delta}} = diag([1, 1, 1])$. This outperformed all other approaches, including PAC-NMPC with the fixed-wing value function (Fig. \ref{fig:fixedwing}). We hypothesize that this may be because the bicycle value function is easier to learn due to its smaller state space.

To demonstrate that our approach is more resilient when the dynamics uncertainty does not match the training environment, we shifted the mean of the noise over the body accelerations to $[3.10, 2.10, 2.11]\frac{m}{s^2}$ and $[4.18, 2.84, 4.71]\frac{rad}{s^2}$ and repeated the experiments. Our approach was able to maintain higher success rates with both the fixed-wing value function and the bicycle value function than PAC-NMPC with a quadratic terminal cost or the actor policy (Fig. \ref{fig:fixedwing_noisy}).

%% file: hardware_experiments.tex
\section{Hardware Experiments}

To evaluate our approach on physical hardware, we used a 1/10\textsuperscript{th} scale Traxxas Rally Car platform with a Velodyne Puck LITE LiDAR. We modeled the rally car dynamics as a stochastic bicycle model with Gaussian process noise fit from data. We used the same value function that was trained entirely in the cluttered environment simulation. We also trained a value function with an incorrect wheelbase, $L=0.5$, to demonstrate the utility of our approach even in the presence of model mismatch. 

We generated 20 random environments consisting of a maximum of 8 circular obstacles in an 8 meter by 6 meter space. We discarded environments in which obstacles overlapped or in which no obstacles were blocking the path to the goal. We placed barrels at these random positions in the motion capture system. In the hardware experiment environments, the Velodyne Puck LITE LiDAR has noise and returns range measurements to the walls, which was not simulated in simulation environments. Thus, these environments are, to some extent, outside of the training distribution.

Our approach outperformed the actor policy and never collided with obstacles (Fig. \ref{fig:hardware_rates}). This indicates that utilizing the RL models inside PAC-NMPC provided better robustness to out-of-distribution environments.

When using the learned value function trained with the incorrect wheelbase, but using the correct wheelbase when sampling PAC-NMPC trajectories, our approach still outperformed the actor policy and never collided with obstacles. Thus, our approach was able to safely utilize an RL model trained in the presence of model mismatch.

%% file: discussion.tex
\begin{figure}[t]
\vspace*{0.75em} 
\centering
\includegraphics[trim={45 0 80 0},clip,width=\linewidth]{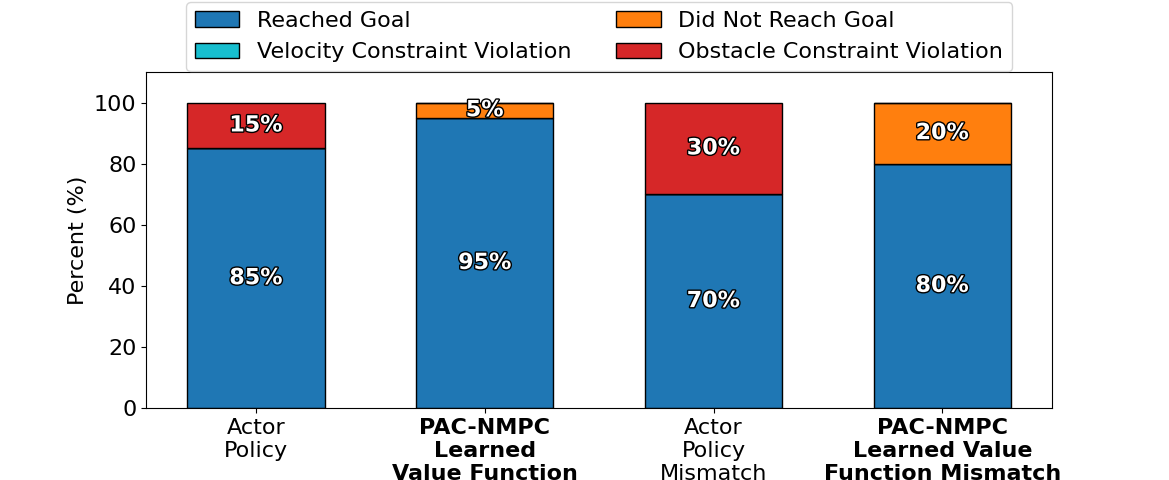}
\caption{Hardware environments. Percentage of trials in which the system reached the goal, didn't reach the goal, and collided.}
\label{fig:hardware_rates}
\vspace{-1.5em} 
\end{figure}

\section{Discussion \& Conclusion}
In this paper, we presented an approach for combining an RL-trained value function with sampling-based SNMPC to achieve probabilistically safe perception-based navigation using only current sensor measurements. We demonstrated, both in simulation and on hardware, that using the learned value function as a terminal cost and constraint enabled the controller to exhibit long-horizon planning while satisfying collision-avoidance constraints. Further, we showed that our approach was able to generate statistical guarantees of performance and safety in real time, which may improve confidence in the controller's ability to use learned components in safety critical environments. We also demonstrated that our approach can scale to complex, high-dimensional systems and that value functions trained on simple dynamics can aid in the control of more complex systems.

Our approach included several limiting assumptions. It assumed the ability to adequately predict future sensor measurements at the terminal NMPC state as well as an accurate stochastic dynamics model. Additionally, while Monte Carlo dropout was able to quantify epistemic model uncertainty, it would be unable to quantify aleatoric uncertainty (i.e., from sensor noise). Our approach also can be affected by uncalibrated uncertainty estimates. Future work could explore the incorporation of perception uncertainty, as well as online refinement of the learned value function.

%% file: main.bbl
\begin{thebibliography}{10}
\providecommand{\url}[1]{#1}
\csname url@rmstyle\endcsname
\providecommand{\newblock}{\relax}
\providecommand{\bibinfo}[2]{#2}
\providecommand\BIBentrySTDinterwordspacing{\spaceskip=0pt\relax}
\providecommand\BIBentryALTinterwordstretchfactor{4}
\providecommand\BIBentryALTinterwordspacing{\spaceskip=\fontdimen2\font plus
\BIBentryALTinterwordstretchfactor\fontdimen3\font minus \fontdimen4\font\relax}
\providecommand\BIBforeignlanguage[2]{{%
\expandafter\ifx\csname l@#1\endcsname\relax
\typeout{** WARNING: IEEEtran.bst: No hyphenation pattern has been}%
\typeout{** loaded for the language `#1'. Using the pattern for}%
\typeout{** the default language instead.}%
\else
\language=\csname l@#1\endcsname
\fi
#2}}

\bibitem{falanga2018pampc}
\BIBentryALTinterwordspacing
D.~Falanga, P.~Foehn, P.~Lu, and D.~Scaramuzza, ``\href{https://ieeexplore.ieee.org/abstract/document/8593739}{PAMPC: Perception-aware model predictive control for quadrotors},'' in \emph{2018 IEEE/RSJ International Conference on Intelligent Robots and Systems (IROS)}.\hskip 1em plus 0.5em minus 0.4em\relax IEEE, 2018, pp. 1--8. [Online]. Available: \url{https://ieeexplore.ieee.org/abstract/document/8593739}
\BIBentrySTDinterwordspacing

\bibitem{brito2019model}
\BIBentryALTinterwordspacing
B.~Brito, B.~Floor, L.~Ferranti, and J.~Alonso-Mora, ``\href{https://ieeexplore.ieee.org/abstract/document/8768044}{Model predictive contouring control for collision avoidance in unstructured dynamic environments},'' \emph{IEEE Robotics and Automation Letters}, vol.~4, no.~4, pp. 4459--4466, 2019. [Online]. Available: \url{https://ieeexplore.ieee.org/abstract/document/8768044}
\BIBentrySTDinterwordspacing

\bibitem{polevoy2022post}
\BIBentryALTinterwordspacing
A.~Polevoy, M.~Basescu, L.~Scheuer, and J.~Moore, ``\href{https://ieeexplore.ieee.org/abstract/document/9812099}{Post-stall navigation with fixed-wing UAVs using onboard vision},'' in \emph{2022 International Conference on Robotics and Automation (ICRA)}.\hskip 1em plus 0.5em minus 0.4em\relax IEEE, 2022, pp. 9696--9702. [Online]. Available: \url{https://ieeexplore.ieee.org/abstract/document/9812099}
\BIBentrySTDinterwordspacing

\bibitem{polevoy2022complex}
\BIBentryALTinterwordspacing
A.~Polevoy, C.~Knuth, K.~M. Popek, and K.~D. Katyal, ``\href{https://ieeexplore.ieee.org/abstract/document/9811644}{Complex terrain navigation via model error prediction},'' in \emph{2022 International Conference on Robotics and Automation (ICRA)}.\hskip 1em plus 0.5em minus 0.4em\relax IEEE, 2022, pp. 9411--9417. [Online]. Available: \url{https://ieeexplore.ieee.org/abstract/document/9811644}
\BIBentrySTDinterwordspacing

\bibitem{lew2021sampling}
T.~Lew and M.~Pavone, ``Sampling-based reachability analysis: A random set theory approach with adversarial sampling,'' in \emph{Conference on robot learning}.\hskip 1em plus 0.5em minus 0.4em\relax PMLR, 2021, pp. 2055--2070.

\bibitem{yin2023risk}
J.~Yin, Z.~Zhang, and P.~Tsiotras, ``Risk-aware model predictive path integral control using conditional value-at-risk,'' in \emph{2023 IEEE International Conference on Robotics and Automation (ICRA)}.\hskip 1em plus 0.5em minus 0.4em\relax IEEE, 2023, pp. 7937--7943.

\bibitem{10250934}
\BIBentryALTinterwordspacing
A.~Polevoy, M.~Kobilarov, and J.~Moore, ``\href{https://ieeexplore.ieee.org/abstract/document/10250934}{Probably Approximately Correct Nonlinear Model Predictive Control (PAC-NMPC)},'' \emph{IEEE Robotics and Automation Letters}, pp. 1--8, 2023. [Online]. Available: \url{https://ieeexplore.ieee.org/abstract/document/10250934}
\BIBentrySTDinterwordspacing

\bibitem{jian2023dynamic}
\BIBentryALTinterwordspacing
Z.~Jian, Z.~Yan, X.~Lei, Z.~Lu, B.~Lan, X.~Wang, and B.~Liang, ``\href{https://ieeexplore.ieee.org/abstract/document/10160857}{Dynamic control barrier function-based model predictive control to safety-critical obstacle-avoidance of mobile robot},'' in \emph{2023 IEEE International Conference on Robotics and Automation (ICRA)}.\hskip 1em plus 0.5em minus 0.4em\relax IEEE, 2023, pp. 3679--3685. [Online]. Available: \url{https://ieeexplore.ieee.org/abstract/document/10160857}
\BIBentrySTDinterwordspacing

\bibitem{wang2022chase}
\BIBentryALTinterwordspacing
C.~Wang, X.~Chen, C.~Li, R.~Song, Y.~Li, and M.~Q.-H. Meng, ``\href{https://ieeexplore.ieee.org/abstract/document/9709214}{Chase and track: Toward safe and smooth trajectory planning for robotic navigation in dynamic environments},'' \emph{IEEE Transactions on Industrial Electronics}, vol.~70, no.~1, pp. 604--613, 2022. [Online]. Available: \url{https://ieeexplore.ieee.org/abstract/document/9709214}
\BIBentrySTDinterwordspacing

\bibitem{singh2022reinforcement}
B.~Singh, R.~Kumar, and V.~P. Singh, ``Reinforcement learning in robotic applications: a comprehensive survey,'' \emph{Artificial Intelligence Review}, vol.~55, no.~2, pp. 945--990, 2022.

\bibitem{zhu2021deep}
K.~Zhu and T.~Zhang, ``Deep reinforcement learning based mobile robot navigation: A review,'' \emph{Tsinghua Science and Technology}, vol.~26, no.~5, pp. 674--691, 2021.

\bibitem{dong2023review}
L.~Dong, Z.~He, C.~Song, and C.~Sun, ``A review of mobile robot motion planning methods: from classical motion planning workflows to reinforcement learning-based architectures,'' \emph{Journal of Systems Engineering and Electronics}, vol.~34, no.~2, pp. 439--459, 2023.

\bibitem{mirowski2016learning}
P.~Mirowski, R.~Pascanu, F.~Viola, H.~Soyer, A.~J. Ballard, A.~Banino, M.~Denil, R.~Goroshin, L.~Sifre, K.~Kavukcuoglu, \emph{et~al.}, ``Learning to navigate in complex environments,'' \emph{arXiv preprint arXiv:1611.03673}, 2016.

\bibitem{zhu2017target}
Y.~Zhu, R.~Mottaghi, E.~Kolve, J.~J. Lim, A.~Gupta, L.~Fei-Fei, and A.~Farhadi, ``Target-driven visual navigation in indoor scenes using deep reinforcement learning,'' in \emph{2017 IEEE international conference on robotics and automation (ICRA)}.\hskip 1em plus 0.5em minus 0.4em\relax IEEE, 2017, pp. 3357--3364.

\bibitem{long2018towards}
P.~Long, T.~Fan, X.~Liao, W.~Liu, H.~Zhang, and J.~Pan, ``Towards optimally decentralized multi-robot collision avoidance via deep reinforcement learning,'' in \emph{2018 IEEE international conference on robotics and automation (ICRA)}.\hskip 1em plus 0.5em minus 0.4em\relax IEEE, 2018, pp. 6252--6259.

\bibitem{kobilarov2015sample}
\BIBentryALTinterwordspacing
M.~Kobilarov, ``\href{https://proceedings.neurips.cc/paper_files/paper/2015/hash/97d98119037c5b8a9663cb21fb8ebf47-Abstract.html}{Sample complexity bounds for iterative stochastic policy optimization},'' \emph{Advances in Neural Information Processing Systems}, vol.~28, 2015. [Online]. Available: \url{https://proceedings.neurips.cc/paper_files/paper/2015/hash/97d98119037c5b8a9663cb21fb8ebf47-Abstract.html}
\BIBentrySTDinterwordspacing

\bibitem{sharma2012autonomous}
\BIBentryALTinterwordspacing
S.~Sharma and M.~E. Taylor, ``\href{http://www.reflexxes.ws/iros2012ws/Paper_12.pdf}{Autonomous waypoint generation strategy for on-line navigation in unknown environments},'' \emph{environment}, vol.~2, p.~3D, 2012. [Online]. Available: \url{http://www.reflexxes.ws/iros2012ws/Paper_12.pdf}
\BIBentrySTDinterwordspacing

\bibitem{kaufmann2018deep}
\BIBentryALTinterwordspacing
E.~Kaufmann, A.~Loquercio, R.~Ranftl, A.~Dosovitskiy, V.~Koltun, and D.~Scaramuzza, ``\href{https://proceedings.mlr.press/v87/kaufmann18a.html}{Deep drone racing: Learning agile flight in dynamic environments},'' in \emph{Conference on Robot Learning}.\hskip 1em plus 0.5em minus 0.4em\relax PMLR, 2018, pp. 133--145. [Online]. Available: \url{https://proceedings.mlr.press/v87/kaufmann18a.html}
\BIBentrySTDinterwordspacing

\bibitem{greatwood2019reinforcement}
\BIBentryALTinterwordspacing
C.~Greatwood and A.~G. Richards, ``\href{https://link.springer.com/article/10.1007/s10514-019-09829-4}{Reinforcement learning and model predictive control for robust embedded quadrotor guidance and control},'' \emph{Autonomous Robots}, vol.~43, pp. 1681--1693, 2019. [Online]. Available: \url{https://link.springer.com/article/10.1007/s10514-019-09829-4}
\BIBentrySTDinterwordspacing

\bibitem{bansal2020combining}
\BIBentryALTinterwordspacing
S.~Bansal, V.~Tolani, S.~Gupta, J.~Malik, and C.~Tomlin, ``\href{https://proceedings.mlr.press/v100/bansal20a}{Combining optimal control and learning for visual navigation in novel environments},'' in \emph{Conference on Robot Learning}.\hskip 1em plus 0.5em minus 0.4em\relax PMLR, 2020, pp. 420--429. [Online]. Available: \url{https://proceedings.mlr.press/v100/bansal20a}
\BIBentrySTDinterwordspacing

\bibitem{brito2021go}
\BIBentryALTinterwordspacing
B.~Brito, M.~Everett, J.~P. How, and J.~Alonso-Mora, ``\href{https://ieeexplore.ieee.org/abstract/document/9385847}{Where to go next: Learning a subgoal recommendation policy for navigation in dynamic environments},'' \emph{IEEE Robotics and Automation Letters}, vol.~6, no.~3, pp. 4616--4623, 2021. [Online]. Available: \url{https://ieeexplore.ieee.org/abstract/document/9385847}
\BIBentrySTDinterwordspacing

\bibitem{bektacs2022apf}
K.~Bekta{\c{s}} and H.~I. Bozma, ``Apf-rl: Safe mapless navigation in unknown environments,'' in \emph{2022 International Conference on Robotics and Automation (ICRA)}.\hskip 1em plus 0.5em minus 0.4em\relax IEEE, 2022, pp. 7299--7305.

\bibitem{gao2023efficient}
Y.~Gao, J.~Wu, X.~Yang, and Z.~Ji, ``Efficient hierarchical reinforcement learning for mapless navigation with predictive neighbouring space scoring,'' \emph{IEEE Transactions on Automation Science and Engineering}, 2023.

\bibitem{chen1998quasi}
\BIBentryALTinterwordspacing
H.~Chen and F.~Allg{\"o}wer, ``\href{https://www.sciencedirect.com/science/article/pii/S0005109898000739}{A quasi-infinite horizon nonlinear model predictive control scheme with guaranteed stability},'' \emph{Automatica}, vol.~34, no.~10, pp. 1205--1217, 1998. [Online]. Available: \url{https://www.sciencedirect.com/science/article/pii/S0005109898000739}
\BIBentrySTDinterwordspacing

\bibitem{zhong2013value}
\BIBentryALTinterwordspacing
M.~Zhong, M.~Johnson, Y.~Tassa, T.~Erez, and E.~Todorov, ``\href{https://ieeexplore.ieee.org/abstract/document/6614995}{Value function approximation and model predictive control},'' in \emph{2013 IEEE symposium on adaptive dynamic programming and reinforcement learning (ADPRL)}.\hskip 1em plus 0.5em minus 0.4em\relax IEEE, 2013, pp. 100--107. [Online]. Available: \url{https://ieeexplore.ieee.org/abstract/document/6614995}
\BIBentrySTDinterwordspacing

\bibitem{lowrey2018plan}
\BIBentryALTinterwordspacing
K.~Lowrey, A.~Rajeswaran, S.~Kakade, E.~Todorov, and I.~Mordatch, ``\href{https://arxiv.org/abs/1811.01848}{Plan online, learn offline: Efficient learning and exploration via model-based control},'' \emph{arXiv preprint arXiv:1811.01848}, 2018. [Online]. Available: \url{https://arxiv.org/abs/1811.01848}
\BIBentrySTDinterwordspacing

\bibitem{hong2019model}
\BIBentryALTinterwordspacing
Z.-W. Hong, J.~Pajarinen, and J.~Peters, ``\href{https://arxiv.org/abs/1908.06012}{Model-based lookahead reinforcement learning},'' \emph{arXiv preprint arXiv:1908.06012}, 2019. [Online]. Available: \url{https://arxiv.org/abs/1908.06012}
\BIBentrySTDinterwordspacing

\bibitem{hoeller2020deep}
\BIBentryALTinterwordspacing
D.~Hoeller, F.~Farshidian, and M.~Hutter, ``\href{https://proceedings.mlr.press/v100/hoeller20a.html}{Deep value model predictive control},'' in \emph{Conference on Robot Learning}.\hskip 1em plus 0.5em minus 0.4em\relax PMLR, 2020, pp. 990--1004. [Online]. Available: \url{https://proceedings.mlr.press/v100/hoeller20a.html}
\BIBentrySTDinterwordspacing

\bibitem{bhardwaj2020blending}
\BIBentryALTinterwordspacing
M.~Bhardwaj, S.~Choudhury, and B.~Boots, ``\href{https://arxiv.org/abs/2012.05909}{Blending mpc \& value function approximation for efficient reinforcement learning},'' \emph{arXiv preprint arXiv:2012.05909}, 2020. [Online]. Available: \url{https://arxiv.org/abs/2012.05909}
\BIBentrySTDinterwordspacing

\bibitem{gu2022review}
S.~Gu, L.~Yang, Y.~Du, G.~Chen, F.~Walter, J.~Wang, and A.~Knoll, ``A review of safe reinforcement learning: Methods, theory and applications,'' \emph{arXiv preprint arXiv:2205.10330}, 2022.

\bibitem{brunke2022safe}
L.~Brunke, M.~Greeff, A.~W. Hall, Z.~Yuan, S.~Zhou, J.~Panerati, and A.~P. Schoellig, ``Safe learning in robotics: From learning-based control to safe reinforcement learning,'' \emph{Annual Review of Control, Robotics, and Autonomous Systems}, vol.~5, no.~1, pp. 411--444, 2022.

\bibitem{geibel2005risk}
P.~Geibel and F.~Wysotzki, ``Risk-sensitive reinforcement learning applied to control under constraints,'' \emph{Journal of Artificial Intelligence Research}, vol.~24, pp. 81--108, 2005.

\bibitem{altman1998constrained}
E.~Altman, ``Constrained markov decision processes with total cost criteria: Lagrangian approach and dual linear program,'' \emph{Mathematical methods of operations research}, vol.~48, pp. 387--417, 1998.

\bibitem{achiam2017constrained}
J.~Achiam, D.~Held, A.~Tamar, and P.~Abbeel, ``Constrained policy optimization,'' in \emph{International conference on machine learning}.\hskip 1em plus 0.5em minus 0.4em\relax PMLR, 2017, pp. 22--31.

\bibitem{bharadhwaj2020conservative}
H.~Bharadhwaj, A.~Kumar, N.~Rhinehart, S.~Levine, F.~Shkurti, and A.~Garg, ``Conservative safety critics for exploration,'' \emph{arXiv preprint arXiv:2010.14497}, 2020.

\bibitem{yu2022towards}
H.~Yu, W.~Xu, and H.~Zhang, ``Towards safe reinforcement learning with a safety editor policy,'' \emph{Advances in Neural Information Processing Systems}, vol.~35, pp. 2608--2621, 2022.

\bibitem{anand2021safe}
A.~Anand, K.~Seel, V.~Gj{\ae}rum, A.~H{\aa}kansson, H.~Robinson, and A.~Saad, ``Safe learning for control using control lyapunov functions and control barrier functions: A review,'' \emph{Procedia Computer Science}, vol. 192, pp. 3987--3997, 2021.

\bibitem{chow2018lyapunov}
Y.~Chow, O.~Nachum, E.~Duenez-Guzman, and M.~Ghavamzadeh, ``A lyapunov-based approach to safe reinforcement learning,'' \emph{Advances in neural information processing systems}, vol.~31, 2018.

\bibitem{fisac2019bridging}
J.~F. Fisac, N.~F. Lugovoy, V.~Rubies-Royo, S.~Ghosh, and C.~J. Tomlin, ``Bridging hamilton-jacobi safety analysis and reinforcement learning,'' in \emph{2019 International Conference on Robotics and Automation (ICRA)}.\hskip 1em plus 0.5em minus 0.4em\relax IEEE, 2019, pp. 8550--8556.

\bibitem{cheng2019end}
R.~Cheng, G.~Orosz, R.~M. Murray, and J.~W. Burdick, ``End-to-end safe reinforcement learning through barrier functions for safety-critical continuous control tasks,'' in \emph{Proceedings of the AAAI conference on artificial intelligence}, vol.~33, no.~01, 2019, pp. 3387--3395.

\bibitem{marvi2021safe}
Z.~Marvi and B.~Kiumarsi, ``Safe reinforcement learning: A control barrier function optimization approach,'' \emph{International Journal of Robust and Nonlinear Control}, vol.~31, no.~6, pp. 1923--1940, 2021.

\bibitem{ma2021model}
H.~Ma, J.~Chen, S.~Eben, Z.~Lin, Y.~Guan, Y.~Ren, and S.~Zheng, ``Model-based constrained reinforcement learning using generalized control barrier function,'' in \emph{2021 IEEE/RSJ International Conference on Intelligent Robots and Systems (IROS)}.\hskip 1em plus 0.5em minus 0.4em\relax IEEE, 2021, pp. 4552--4559.

\bibitem{yang2022safe}
T.-Y. Yang, T.~Zhang, L.~Luu, S.~Ha, J.~Tan, and W.~Yu, ``Safe reinforcement learning for legged locomotion,'' in \emph{2022 IEEE/RSJ International Conference on Intelligent Robots and Systems (IROS)}.\hskip 1em plus 0.5em minus 0.4em\relax IEEE, 2022, pp. 2454--2461.

\bibitem{sheckells2019actor}
\BIBentryALTinterwordspacing
M.~Sheckells, G.~Garimella, S.~Michra, and M.~Kobilarov, ``\href{https://par.nsf.gov/servlets/purl/10136847}{Actor-critic pac robust policy search},'' \emph{ICRA 2019}, 2019. [Online]. Available: \url{https://par.nsf.gov/servlets/purl/10136847}
\BIBentrySTDinterwordspacing

\bibitem{zanon2020safe}
M.~Zanon and S.~Gros, ``Safe reinforcement learning using robust mpc,'' \emph{IEEE Transactions on Automatic Control}, vol.~66, no.~8, pp. 3638--3652, 2020.

\bibitem{gros2022learning}
S.~Gros and M.~Zanon, ``Learning for mpc with stability \& safety guarantees,'' \emph{Automatica}, vol. 146, p. 110598, 2022.

\bibitem{sutton2018reinforcement}
\BIBentryALTinterwordspacing
R.~S. Sutton and A.~G. Barto, \emph{\href{https://www.andrew.cmu.edu/course/10-703/textbook/BartoSutton.pdf}{Reinforcement learning: An introduction}}, 2020. [Online]. Available: \url{https://www.andrew.cmu.edu/course/10-703/textbook/BartoSutton.pdf}
\BIBentrySTDinterwordspacing

\bibitem{tai2017virtual}
\BIBentryALTinterwordspacing
L.~Tai, G.~Paolo, and M.~Liu, ``\href{https://ieeexplore.ieee.org/abstract/document/8202134}{Virtual-to-real deep reinforcement learning: Continuous control of mobile robots for mapless navigation},'' in \emph{2017 IEEE/RSJ International Conference on Intelligent Robots and Systems (IROS)}.\hskip 1em plus 0.5em minus 0.4em\relax IEEE, 2017, pp. 31--36. [Online]. Available: \url{https://ieeexplore.ieee.org/abstract/document/8202134}
\BIBentrySTDinterwordspacing

\bibitem{de2021soft}
\BIBentryALTinterwordspacing
J.~C. de~Jesus, V.~A. Kich, A.~H. Kolling, R.~B. Grando, M.~A. d. S.~L. Cuadros, and D.~F.~T. Gamarra, ``\href{https://link.springer.com/article/10.1007/s10846-021-01367-5}{Soft actor-critic for navigation of mobile robots},'' \emph{Journal of Intelligent \& Robotic Systems}, vol. 102, no.~2, p.~31, 2021. [Online]. Available: \url{https://link.springer.com/article/10.1007/s10846-021-01367-5}
\BIBentrySTDinterwordspacing

\bibitem{beomsoo2021mobile}
\BIBentryALTinterwordspacing
H.~Beomsoo, A.~A. Ravankar, and T.~Emaru, ``\href{https://ieeexplore.ieee.org/abstract/document/9419565}{Mobile robot navigation based on deep reinforcement learning with 2d-lidar sensor using stochastic approach},'' in \emph{2021 IEEE International Conference on Intelligence and Safety for Robotics (ISR)}.\hskip 1em plus 0.5em minus 0.4em\relax IEEE, 2021, pp. 417--422. [Online]. Available: \url{https://ieeexplore.ieee.org/abstract/document/9419565}
\BIBentrySTDinterwordspacing

\bibitem{huang2022cleanrl}
\BIBentryALTinterwordspacing
S.~Huang, R.~F.~J. Dossa, C.~Ye, J.~Braga, D.~Chakraborty, K.~Mehta, and J.~G. Araújo, ``Cleanrl: High-quality single-file implementations of deep reinforcement learning algorithms,'' \emph{Journal of Machine Learning Research}, vol.~23, no. 274, pp. 1--18, 2022. [Online]. Available: \url{http://jmlr.org/papers/v23/21-1342.html}
\BIBentrySTDinterwordspacing

\bibitem{fujimoto2018addressing}
\BIBentryALTinterwordspacing
S.~Fujimoto, H.~Hoof, and D.~Meger, ``\href{https://proceedings.mlr.press/v80/fujimoto18a.html}{Addressing function approximation error in actor-critic methods},'' in \emph{International conference on machine learning}.\hskip 1em plus 0.5em minus 0.4em\relax PMLR, 2018, pp. 1587--1596. [Online]. Available: \url{https://proceedings.mlr.press/v80/fujimoto18a.html}
\BIBentrySTDinterwordspacing

\bibitem{gal2016dropout}
\BIBentryALTinterwordspacing
Y.~Gal and Z.~Ghahramani, ``\href{https://proceedings.mlr.press/v48/gal16.html?trk=public_post_comment-text}{Dropout as a bayesian approximation: Representing model uncertainty in deep learning},'' in \emph{international conference on machine learning}.\hskip 1em plus 0.5em minus 0.4em\relax PMLR, 2016, pp. 1050--1059. [Online]. Available: \url{https://proceedings.mlr.press/v48/gal16.html?trk=public_post_comment-text}
\BIBentrySTDinterwordspacing

\bibitem{7487277}
G.~Williams, P.~Drews, B.~Goldfain, J.~M. Rehg, and E.~A. Theodorou, ``Aggressive driving with model predictive path integral control,'' in \emph{2016 IEEE International Conference on Robotics and Automation (ICRA)}, 2016, pp. 1433--1440.

\bibitem{basescu2020direct}
M.~Basescu and J.~Moore, ``Direct nmpc for post-stall motion planning with fixed-wing uavs,'' in \emph{2020 IEEE International Conference on Robotics and Automation (ICRA)}.\hskip 1em plus 0.5em minus 0.4em\relax IEEE, 2020, pp. 9592--9598.

\end{thebibliography}
